\documentclass[10pt,twocolumn,letterpaper]{article}

\usepackage{cvpr}              
\definecolor{cvprblue}{rgb}{0.21,0.49,0.74}
\usepackage[pagebackref,breaklinks,colorlinks,allcolors=cvprblue]{hyperref}
\usepackage{graphicx}
\usepackage{subcaption} 
\def\paperID{49} 
\def\confName{CVPR}
\def\confYear{2026}

\title{Representation-Conditioned Diffusion Models for \\ Guided Training Data Generation}

\author{
Nithesh Chandher Karthikeyan, Jonas Unger, Gabriel Eilertsen\\
Linköping University\\
{\tt\small \{nithesh.chandher.karthikeyan, jonas.unger, gabriel.eilertsen\}@liu.se}
}

\begin{document}
\maketitle
\begin{abstract}

Data availability remains a critical bottleneck in many deep learning applications. Large-scale datasets are often expensive to collect, curate and annotate, which can limit the scalability and applicability of supervised learning methods. In this work, we evaluate the classification performance of models trained on synthetic image datasets produced by generative deep learning. In particular, we use latent diffusion models conditioned on learned representations from DINOv2, DINOv3, and CLIP. Our results demonstrates that this representation-conditioned formulation significantly outperforms class-conditioned generation by a large margin (+10.76 p.p. top-1 accuracy on ImageNet100), by improving sample quality and mode coverage. Furthermore, by scaling the size of the synthetic dataset, we are able to outperform a classifier trained on the real data (+2.0 p.p. top-1 accuracy). 

We also demonstrate how generated images can be used for augmentation purposes, outperforming classical augmentation methods, and how the conditioning space can be used for sample filtering to further improve training value. 
Collectively, these findings highlight that representation-conditioned diffusion models provide a promising approach for augmenting, complementing, or potentially replacing real-world datasets in large-scale visual learning tasks.

\end{abstract}    
\section{Introduction}
\label{sec:intro}

\begin{figure*}[t]
\includegraphics[width=\linewidth]{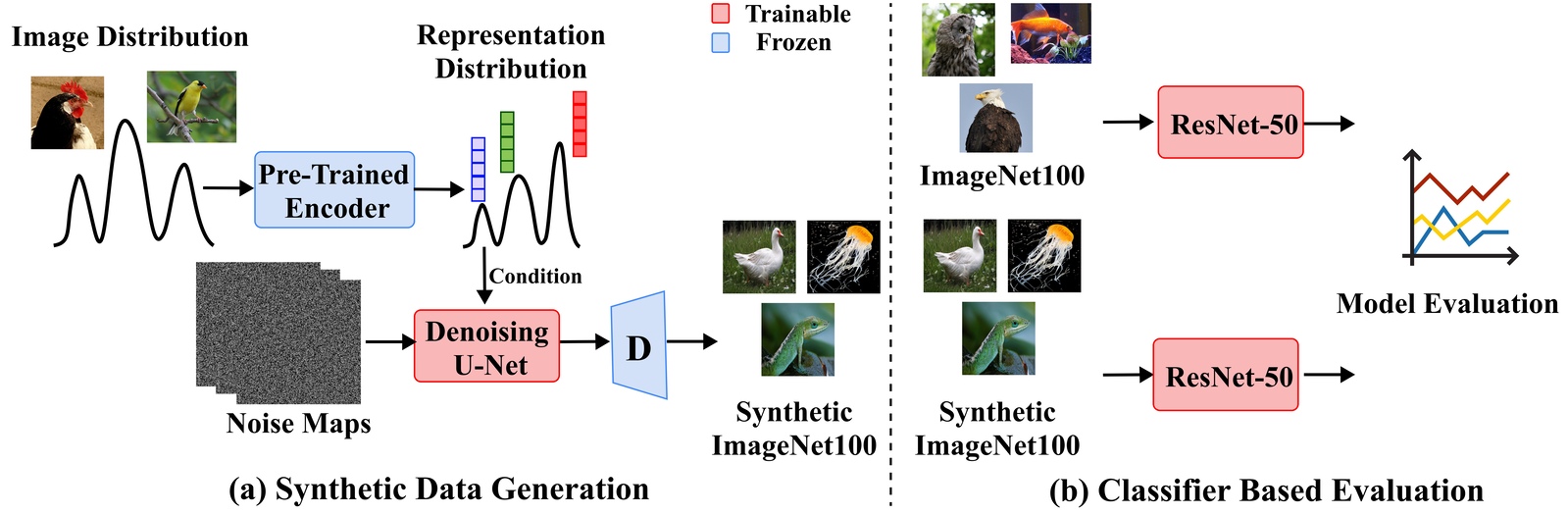}
\caption{Overall pipeline of the methodology, which consist of two parts: (a) Synthetic Data Generation using Representation Conditioned Diffusion Model (RCDM) and (b) Classifier Based Evaluation for Comparative analysis.}
\label{fig:pipeline}
\end{figure*}

With continuous advancements in deep learning, there is a constant demand for larger and diverse training datasets. At the same time, generative modeling has experienced significant progress over the last decade, particularly with the emergence of diffusion models~\cite{sohl2015deep,ho2020denoising}. These developments have sparked a growing interest in using these models to expand training datasets via synthetic data generation, e.g., by steering or fine-tuning a general pre-trained text-to-image model~\cite{azizi2023synthetic,he2023is,trabucco2024effective}.


In this paper, we focus on the most challenging scenario for generative dataset synthesis. Specifically, we assume that (1) only the target dataset is available, preventing the use of existing text-to-image models, (2) the downstream model is trained exclusively on the generated dataset (without access to real images), and (3) no additional annotations, such as text prompts, are available to guide generation. Unlike conventional data augmentation, where augmented images only introduce variations to the real samples, training solely on synthetic dataset requires the generative model to capture the full diversity of the real data distribution. Further, this scenario is also relevant in situations where sharing of real data are restricted due to privacy concerns, such as in medical imaging. 




Text-to-image diffusion models can produce high quality and diverse samples, with their performance largely benefiting from guidance provided by text prompts, in addition to large models and training dataset. To increase the quality and diversity without text guidance, we consider representation-conditioned diffusion models (RCDMs)~\cite{bordes2021high,li2024return,jimenez2025dino}. In RCDMs, guidance is provided from extracted representations of each image, which has been demonstrated to significantly increase the quality of generation as compared to unconditioned or class-conditioned generation~\cite{li2024return}. In this work, we evaluate how these improvements translate to the case of training data generation.


We provide experiments with a representation-conditioned latent diffusion model~\cite{rombach2022high} trained on ImageNet100, which is a challenging dataset due to its diversity and relatively limited size. We demonstrate that RCDMs show great promise for synthetic dataset generation, surpassing the class-conditioned generation by a large margin. Furthermore, by scaling the size of the synthetic dataset, we are able to outperform the model trained only on the real dataset. We see this as quite remarkable evidence for the promise of RCDMs and using generative deep learning for training data synthesis.

\vspace{2mm}
\noindent In summary, we present the following contributions:
\begin{itemize}
        \item We evaluate the classification performance of models trained on synthetic datasets generated by RCDMs, investigating the impact of dataset size and representation used for conditioning (DINOv2~\cite{oquab2023dinov2}, DINOv3~\cite{simeoni2025dinov3}, and CLIP~\cite{radford2021learning}).
        
        \item We show that, by scaling the synthetic dataset, RCDMs can surpass the performance of models trained solely on real data, highlighting their potential for large-scale dataset generation.
    
        \item We further demonstrate that RCDM generated images are effective for data augmentation, outperforming classical techniques such as RandAugment~\cite{cubuk2020randaugment} and MixUp~\cite{zhang2017mixup}, and how the representation space used for conditioning allows simple sample filtering.
\end{itemize}

\vspace{2mm}
\noindent While this work demonstrates a first case study of using RCDMs for dataset generation, the results open up for many possibilities in future research, for improving or replacing datasets by means of generative synthesis.
\section{Method}
\label{sec:method}

In this section, we describe the pipeline used for generation and validation of synthetic training data. The pipeline consists of two main stages: synthetic data generation using Representation Conditioned Diffusion Models (RCDMs) and evaluation using a classifier.  Figure~\ref{fig:pipeline} illustrates the overall workflow, which is described in detail in the following subsections.

\subsection{Synthetic Data Generation using RCDM}
Training the RCDM involves two stages. First, input images are projected into a lower-dimensional representation space using a pre-trained encoder. Second, a latent diffusion model (LDM)~\cite{rombach2022high} is trained using these representations as conditioning signals~\cite{karthikeyan5772685evaluating}.

For the experiments, we train three RCDM models conditioned on representations extracted from the image encoder of CLIP~\cite{radford2021learning} (also called unCLIP~\cite{ramesh2022hierarchical}), DINOv2~\cite{oquab2023dinov2}, and DINOv3~\cite{simeoni2025dinov3}. Specifically, we use the CLS token embeddings from the DINOv2 ViT-B/14 and DINOv3 ViT-B/16 architectures, each producing 768-dimensional representations, and embeddings from CLIP ViT-B/32 with a dimensionality of 512. Training is performed on the ImageNet-100 dataset, a subset of ImageNet containing 100 classes~\cite{russakovsky2015imagenet}. The dataset contains 130k images, all resized to a resolution of 256×256. Each model was trained for 125 epochs with a batch size of 16, resulting in approximately 1.16 million training steps.

For sampling, we use the Denoising Diffusion Probabilistic Model (DDPM)~\cite{ho2020denoising} with 100 inference steps. To increase dataset diversity beyond the original 130k images, images are generated using different random seeds, each corresponding to a distinct latent noise initialization. This produces diverse images while preserving semantic content through representation conditioning~\cite{karthikeyan5772685evaluating}.

\subsection{Classifier Based Evaluation}
To evaluate the quality of the synthetic ImageNet-100 data in relation to the real dataset, we train separate classifiers on the real and synthetic datasets using the ResNet-50~\cite{he2016deep} architecture. This evaluation measures how effectively models trained on synthetic data can learn discriminative features comparable to those learned from real data.

The training procedure follows the standard configuration used for training ResNet-50 on ImageNet in the ImageNet Large Scale Visual Recognition Challenge (ILSVRC). Specifically, the classifier is trained using the same architectural setup, data preprocessing pipeline, and optimization strategy adopted in ILSVRC. By maintaining an identical training configuration across both datasets, the comparison provides a consistent and fair assessment of the representational quality and class-discriminative structure present in the generated synthetic images. 
\label{sec:result}
\begin{figure*}[t]
\includegraphics[width=\textwidth]{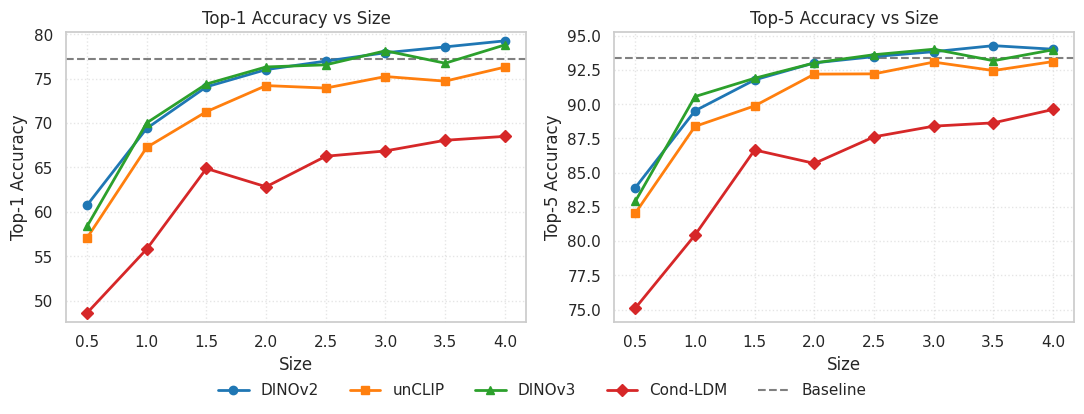}
\caption{Scaling synthetic datasets for classifier training. Top-1 (left) and Top-5 (right) accuracy of a ResNet-50 trained on synthetic ImageNet-100 datasets generated using representation-conditioned diffusion models with DINOv2, DINOv3, and CLIP representations, compared to a class-conditioned LDM baseline and ResNet-50 trained on real ImageNet100 baseline.}
\label{fig:result}
\end{figure*}

\section{Results}

This section presents the experimental evaluation of the proposed framework. We assess the quality and usefulness of the generated synthetic images using quantitative downstream classification experiments on the ImageNet-100 dataset. 

\subsection{Synthetic Dataset Scaling} 
In this experiment, we investigate whether the performance gap between classifiers trained on real and synthetic data can be reduced using synthetic datasets generated by RCDMs, and further analyze this by scaling the synthetic dataset beyond the size of the real dataset to evaluate the impact of increased scale and diversity on downstream classification performance.

For this evaluation, we train classifiers based on the ResNet50 architecture under multiple training datasets. The baseline model is trained on the real ImageNet100 dataset, which achieves a Top-1 accuracy of $77.26\%$ and a Top-5 accuracy of $93.38\%$. In addition, three classifiers are trained on synthetic datasets generated using RCDMs conditioned on representations extracted from DINOv2, DINOv3, and CLIP. As an additional baseline for synthetic data generation, we include a class-conditioned latent diffusion model (Cond-LDM~\cite{rombach2022high}), which generates images conditioned solely on class labels.

The results shown in Figure~\ref{fig:result} demonstrate that classifiers trained on synthetic datasets generated using representations from DINOv2 and DINOv3 outperform the baseline model trained on real data when the dataset size is increased to three times the size of the real dataset. The best result is obtained using the dataset generated with DINOv2 representations, achieving $79.26\%$ Top-1 accuracy, when the dataset is scaled to four times the size of the real dataset. The dataset generated with DINOv3 representations performs comparably, reaching $78.8\%$ Top-1 accuracy at the same scale. Although the synthetic dataset generated using CLIP representations does not exceed the baseline accuracy, it achieves a competitive Top-1 accuracy of $76.3\%$ at four times the size of the real dataset and has the potential to surpass the baseline score based on the observed trend.

Another notable observation is the superior performance of datasets generated using RCDMs compared to the class-conditioned diffusion baseline. The classifier trained on data generated by Cond-LDM achieves a highest Top-1 accuracy of $68.5\%$, leaving a significant gap relative to the baseline model trained on real data. This highlights the importance of rich semantic conditioning via pre-trained vision encoders, which enables the diffusion model to better capture the underlying structure of the data distribution.

\subsection{Filtering in Representation Space}
To study the effect of filtering in representation space, we first generate synthetic datasets of twice the size of the real dataset using RCDMs conditioned on DINOv2, DINOv3, and CLIP. The generated images are then mapped back to the representation space, where class centroids are computed. For each class, the $25\%$ nearest and $25\%$ farthest representations are removed, retaining only the middle samples. The corresponding images form a filtered dataset with the same size as the real dataset. As shown in Table~\ref{tab:hem_results}, this filtering achieves slight improvements over using unfiltered synthetic data, although a noticeable gap to the real-data baseline remains. Nevertheless, this result suggests that representation-space filtering can be a promising sampling strategy, which is difficult to attain in other generative frameworks such as text-to-image models.

\begin{table}[t]
\centering
\small
\setlength{\tabcolsep}{4pt}
\caption{Classification performance comparison on ImageNet100 with different synthetic datasets and sample filtering strategies.}
\label{tab:hem_results}
\begin{tabular}{lccc}
\toprule
\textbf{Experiment} & \textbf{Size} & \textbf{Top-1 (\%)} & \textbf{Top-5 (\%)} \\
\midrule
Baseline & 1.0 & 77.26 & 93.38 \\
\midrule
DINOv2-ResNet50 & 1.0 & 69.40 & 89.52 \\
DINOv3-ResNet50 & 1.0 & 70.02 & 90.56 \\
CLIP-ResNet50 & 1.0 & 67.26 & 88.38 \\
\midrule
Sample Filtering (DINOv2) & 1.0 & 69.98 & 90.14 \\
\textbf{Sample Filtering (DINOv3)} & \textbf{1.0} & \textbf{71.10} & \textbf{90.10} \\
Sample Filtering (CLIP) & 1.0 & 66.22 & 88.16 \\
\bottomrule
\end{tabular}
\end{table}

\subsection{Data Augmentation vs Synthetic Data}

To further assess the effectiveness of the generated synthetic datasets for data augmentation, we compare their performance with traditional data augmentation techniques. Along with the baseline, we evaluate classifiers trained using common augmentation strategies including AutoAug~\cite{cubuk2018autoaugment}, RandAug~\cite{cubuk2020randaugment}, Mixup~\cite{zhang2017mixup}, and ME-ADA~\cite{zhao2020maximum}. These methods are known to improve generalization by increasing the diversity of the training data through transformations applied directly to real images.

As shown in Table~\ref{tab:augmentation_results}, traditional augmentation techniques significantly improve the performance of the baseline model. In particular, Mixup achieves the highest Top-1 accuracy of $81.58\%$, followed closely by AutoAug with $81.28\%$. RandAug and ME-ADA also provide improvements over the baseline, reaching $79.88\%$ and $80.14\%$ Top-1 accuracy, respectively.

Classifiers trained on real data augmented with synthetic images generated by RCDMs achieve comparable or superior performance to standard augmentation methods. In particular, adding synthetic data generated with DINOv2 representations (twice the size of real dataset, marked as x2 in Table~\ref{tab:augmentation_results}) achieves the best result with $82.22\%$ Top-1 and $95.86\%$ Top-5 accuracy, outperforming all evaluated augmentation strategies. Synthetic data generated using DINOv3 and CLIP representations also achieve competitive results, highlighting the effectiveness of representation-conditioned diffusion for providing an alternative to conventional data augmentation methods.

\begin{table}[t]
\centering
\small
\caption{Classification performance comparison on ImageNet100 using different data augmentation methods and synthetic datasets generated by representation conditioned diffusion models.}
\label{tab:augmentation_results}
\begin{tabular}{lccc}
\toprule
\textbf{Experiment} & \textbf{Size} & \textbf{Top-1 (\%)} & \textbf{Top-5 (\%)} \\
\midrule
Baseline & 1.0 & 77.26 & 93.38 \\
AutoAug & 1.0 & 81.28 & 95.54 \\
RandAug & 1.0 & 79.88 & 95.14 \\
Mixup & 1.0 & 81.58 & 95.54 \\
ME-ADA & 1.0 & 80.14 & 94.62 \\
\midrule
Baseline + DINOv2 & 2.0 & 81.90 & 95.80 \\
Baseline + DINOv3 & 2.0 & 81.08 & 95.36 \\
Baseline + CLIP & 2.0 & 81.78 & 95.30 \\
Baseline + Cond-LDM & 2.0 & 79.62 & 94.58 \\
\midrule
\textbf{Baseline + x2 DINOv2} & \textbf{3.0} & \textbf{82.22} & \textbf{95.86} \\
Baseline + x2 DINOv3 & 3.0 & 81.70 & 95.76 \\
Baseline + x2 CLIP & 3.0 & 80.74 & 95 \\
Baseline + x2 Cond-LDM & 3.0 & 80.68 & 94.94 \\
\bottomrule
\end{tabular}
\end{table}
\section{Discussion and Conclusion}
\label{sec:conclusion}

We have demonstrated how RCDMs can pose as a powerful training data generation engine, which can be used to produce purely synthetic datasets or for augmentation purposes. The representation conditioning provides an effective self-supervised guidance mechanism for increasing quality and diversity, similar to how text prompts guide text-to-image models, and we have demonstrated that this improvement translates to improved training data value. 
In particular, representations from DINO-based encoders provide strong semantic conditioning signals that help reduce the domain gap between real and synthetic data and significantly improve the usefulness of generated datasets for downstream classification tasks.
By scaling the size of a purely synthetic ImageNet100 dataset we are able to outperform training on the real images, which is a promising achievement for generative training data synthesis considering the complexity of the dataset. Despite the recent progress in generative deep learning, until now it has been difficult to reach similar or better performance compared to training on real data. For example, a class-conditioned diffusion model is still far behind, and representation conditioning provides a promising strategy for increasing training value.

We see many possible directions for future research using RCDMs for training data generation. Most centrally, there are many possibilities in improving the sampling of RCDMs through the representation space used for conditioning. For example, it is possible to transform representations to further improve training value, e.g., by performing interpolation, hard example mining or MixUp in this space. Also, while in general RCDMs provide high sample quality, there are still some low-quality generations such that a mechanism for removing outliers would likely improve the synthetic dataset. Additionally, further experiments are required for verifying the performance of RCDMs for training data generation on other datasets and under different data availability scenarios.
{
    \small
    \bibliographystyle{ieeenat_fullname}
    \bibliography{main}
}


\end{document}